\documentclass[12pt]{spieman}  
\usepackage{amsmath,amsfonts,amssymb}
\usepackage{graphicx}
\usepackage{setspace}
\usepackage{tocloft}
\usepackage{lineno}

\usepackage{framed,multirow}

\usepackage{amssymb}
\usepackage{latexsym}
\usepackage{siunitx}

\usepackage{multirow}
\usepackage{graphicx}
\usepackage{booktabs}
\usepackage{caption}
\usepackage{array}
\usepackage{makecell}

\usepackage{url}
\usepackage{xcolor}

\usepackage{pifont}
\newcommand{\cmark}{\ding{51}}%
\newcommand{\xmark}{\ding{55}}%

\usepackage[ruled,vlined,algo2e]{algorithm2e}
\SetKwComment{Comment}{$\triangleright$\ }{}

\usepackage{mathtools}
\usepackage{amsmath}

\usepackage{algorithm}
\usepackage{dblfloatfix}
\usepackage[numbers]{natbib}

\title{SNR-ST-Mix: Sample-specific Neighborhood Regression Mixup for Augmented Spatial Transcriptomics Imputation with Deep Neural Network}

\author[a, b]{Hongyi Yu}
\author[a]{Yaoyu Fang}
\author[a]{Jiahe Qian}
\author[c]{Xinkun Wang}
\author[d]{Lee A. Cooper}
\author[a]{Bo Zhou}
\affil[a]{Northwestern University, Department of Radiology, Chicago, IL, USA}
\affil[b]{Yale University, Department of Biostatistics, New Haven, CT, USA}
\affil[c]{Northwestern University, Department of Cell and Developmental Biology, Chicago, IL, USA}
\affil[d]{Northwestern University, Department of Pathology, Chicago, IL, USA}

\cftpagenumbersoff{figure}
\cftpagenumbersoff{table} 
\begin{document} 
\maketitle

\begin{abstract}

\textbf{Purpose:}
Spatial transcriptomics (ST) enables gene expression measurements within the tissue context. However, these measurements are often noisy, low-resolution, and sparsely sampled, which limits the recovery of fine spatial structure. Deep neural networks have become powerful tools for expression imputation from histology, but their performance remains constrained by limited sample sizes and a lack of biologically informed augmentation. Most of the existing augmentation strategies for learning are designed for classification tasks rather than regression, which neglect spatial and transcriptomic relationships, leading to biologically implausible interpolations that hinder prediction performance. 

\textbf{Approach:}
To address these limitations, we propose SNR-ST-Mix, a geometry- and expression-aware data augmentation framework designed specifically for ST data. It constrains mixing to a spot’s k-nearest spatial neighbors and adaptively weights interpolation coefficients based on expression similarity, generating augmented samples that preserve local biological structure while ensuring spatial smoothness. This dual conditioning yields synthetic examples that expand the effective training manifold, promote generalization, and enhance prediction stability under sample-specific training. 

\textbf{Results:}
Extensive experiments with various tissue types demonstrate that SNR-ST-Mix consistently outperforms conventional augmentation methods without requiring architectural changes or additional computation. 

\textbf{Conclusions:}
SNR-ST-Mix provides an effective and biologically principled augmentation strategy for spatial transcriptomics regression tasks. By explicitly leveraging spatial geometry and transcriptomic similarity, it expands the effective training manifold and improves predictive performance without increasing model complexity.

\end{abstract}


{\noindent \footnotesize\textbf{*}Hongyi Yu,  
\linkable{hongyi.yu@yale.edu} }

{\noindent \footnotesize\textbf{*}Bo Zhou,  \linkable{bo.zhou@northwestern.edu} }

\begin{spacing}{1}

\section{Introduction}

Spatial transcriptomics (ST) has emerged as a transformative technology that enables the measurement of gene expression profiles while preserving spatial context within intact tissue sections \cite{Marx2021}. By mapping molecular signals onto histological morphology, ST integrates transcriptional and structural information to provide unprecedented insight into the organization of complex multicellular systems \citep{jiang2023, asp2020}. This integration has become central to modern biomedical research, allowing molecular programs to be localized within specific anatomical niches. In oncology, for instance, combining ST with single-cell RNA sequencing reveals intratumoral heterogeneity and cellular interactions that drive disease progression \citep{Arora2023, Cilento2024, Ji2020}. In cardiology, spatially resolved transcriptomic profiling across infarct border zones elucidates time-dependent molecular remodeling after myocardial infarction \citep{Kuppe2022, Kanemaru2023}. Beyond cancer and cardiology, ST has shown wide applicability in neuroscience, immunology, and developmental biology, supporting spatially variable gene discovery, biomarker identification, and disease stratification while preserving positional context \citep{Jain2024, Wang2023}.

Despite its power, ST remains limited by high cost, low throughput, and workflow complexity. Slide capture, library preparation, and deep sequencing each contribute substantial per-sample expense, and specialized instruments restrict scalability for large studies or clinical translation \citep{Choe2023, Smith2024, Ruiz2025}. These challenges are amplified when multiple time points or tissue conditions must be profiled, making comprehensive spatial atlases prohibitively expensive and logistically difficult to standardize across laboratories \citep{Jones2024Optimizing, Juwayria2024MIST, Piwecka2023NRN}. Furthermore, technical factors such as variable tissue integrity, uneven capture efficiency, and differences in sequencing depth can compromise data quality, introducing noise and dropout events that obscure spatial structure and reduce reproducibility \citep{You2024SystematicST, Liu2022SPCS}.

As a result, many ST datasets remain sparse, noisy, and incomplete, limiting the ability to recover sharp expression boundaries, identify spatial domains, or compare patterns across samples \citep{Zhao2024SuperResST}. These challenges have motivated a growing interest in computational expression imputation, in which molecular information is inferred or refined using histology images \citep{He2020, Schmauch2020}. Because histology slides are ubiquitous, low-cost, and already integrated into diagnostic pipelines, histology-guided imputation offers a scalable complement to sequencing-based ST by enhancing signal quality, increasing effective resolution, and enabling the construction of high-throughput virtual spatial transcriptomic maps \citep{Li2024, Alsaafin2023}. Moreover, histology-based computation can facilitate longitudinal studies, multi-site collaborations, and the inclusion of rare or archival samples, empowering a new wave of translational and population-scale research \citep{Pang2021, Xie2023, Chen2024}.

Deep learning has become the dominant strategy for this task, offering the representational capacity required to link complex histomorphological patterns with underlying gene expression. Early models such as HE2RNA demonstrated that histopathology encodes sufficient signal to reconstruct bulk RNA profiles from whole-slide images (WSIs), motivating subsequent spatially supervised frameworks for spot-level prediction \citep{Schmauch2020}. STNet first trained convolutional networks to regress expression values at each Visium spot \citep{He2020}, followed by HisToGene \citep{Pang2021} and Hist2ST \citep{Zeng2022}, which formulated multi-task regression problems and incorporated spatial dependencies among neighboring regions to capture tissue organization. More recent methods, such as BLEEP \citep{Xie2023} and HGGEP \citep{Li2024}, introduced multimodal contrastive learning and transformer-based architectures to improve cross-modal feature alignment and global context reasoning. However, these advances face a fundamental limitation: paired H\&E–ST datasets remain small, highly tissue-specific, and heterogeneous in data quality. Even when large public repositories exist, variations in staining, scanner calibration, or section thickness cause domain shifts that degrade cross-cohort performance. Models trained on external cohorts often exhibit strong in-sample accuracy but generalize poorly to unseen tissues, highlighting a lack of robustness and reproducibility \citep{Stacke2021DomainShift, Sikaroudi2023OOD}. To mitigate these challenges, sample-specific training strategies have been proposed. For example, S2S-ST employs a customized cascade data-consistent imputation network to reconstruct sample-specific dense gene expression maps from sparse measurements \citep{Fang2025}. ST-DAI performs intra-sample domain adaptation to generate pseudo-supervision within a target sample \citep{Qian2025}. These approaches enhance in-distribution fidelity by maintaining strict alignment with the target tissue, but they inherently limit scalability and knowledge transfer across samples, as each model learns from only a single or few examples, reducing overall diversity and generalization power.

Data augmentation provides a simple yet powerful means to overcome these data scarcity issues and improve the generalization of sample-specific models \citep{Jin2024}. Among various augmentation techniques, mixup (which interpolates both samples and labels) has proven particularly effective for improving robustness, smoothing decision boundaries, and regularizing over-parameterized neural networks \citep{Zhang2017, Yu2021MixupWithoutHesitation}. Advanced variants such as CutMix \citep{Yun2019} and PuzzleMix \citep{Kim2020} manipulate image regions and combine discrete class labels to expand the effective training distribution. However, these methods are tailored for classification tasks and assume categorical targets, making them unsuitable for continuous regression objectives such as gene expression prediction. Naive interpolation in pixel or label space can generate biologically implausible or noisy gene expression targets, leading to unstable optimization. Regression-aware variants like C-Mixup \citep{Yao2022} and RC-Mixup \citep{Hwang2024} partially address this by enforcing label similarity constraints or integrating robust loss functions, but they remain designed for generic tabular or image regression tasks and do not exploit the unique spatial structure inherent to ST data. In ST, neighboring tissue regions are not independent but rather exhibit correlated morphology and expression gradients reflective of microenvironmental continuity. Neglecting this structure disrupts biological coherence in augmented samples, introducing artificial spatial discontinuities that reduce interpretability and degrade prediction fidelity. Thus, there is a pressing need for data augmentation methods that respect both spatial geometry and biological similarity when applied to spatially resolved omics.

To address these challenges, we propose Sample-specific Neighborhood Regression ST Mixup (SNR-ST-Mix), a geometry- and expression-aware data augmentation framework specifically tailored for ST. Unlike conventional mixup that combines random pairs of samples, SNR-ST-Mix introduces dual conditioning based on spatial proximity and expression similarity to generate biologically coherent interpolations. Each spot is augmented only with its $k$ nearest spatial neighbors within the tissue topology, preserving local anatomical context and microstructural organization. Meanwhile, mixing probabilities are adaptively weighted by a Gaussian kernel over gene expression distances, ensuring that augmented examples derive from transcriptionally similar regions. This design yields convex combinations that follow biological gradients and reflect realistic transitions across tissue compartments, thus expanding the training manifold without compromising spatial fidelity. By synthesizing new, spatially consistent training examples, SNR-ST-Mix improves data efficiency, mitigates overfitting, and enhances stability during regression learning, particularly under sample-specific training regimes where data availability is inherently constrained. Importantly, the method is architecture-agnostic, computationally lightweight, and easily integrable into existing pipelines, providing a practical and generalizable solution for spatial transcriptomic imputation.

Our main contributions are as follows:

(1) We introduce SNR-ST-Mix, the first regression-based mixup framework specifically designed for spatial transcriptomic data, which enforces both spatial locality and expression similarity to generate biologically meaningful and spatially coherent augmentations.

(2) SNR-ST-Mix bridges externally trained and sample-specific paradigms, enhancing robustness to data scarcity while maintaining strong in-distribution generalization. This effectively mitigates overfitting, improves prediction stability, and reduces dependence on large external datasets.

(3) Our approach requires no additional network modules and integrates seamlessly into standard architectures. We demonstrate that SNR-ST-Mix consistently outperforms conventional augmentations across multiple ST datasets, showing high adaptability across tissue types and experimental conditions.

\section{Methods}
\begin{figure*}[htb] 
    \centering
    \includegraphics[width=0.98\textwidth]{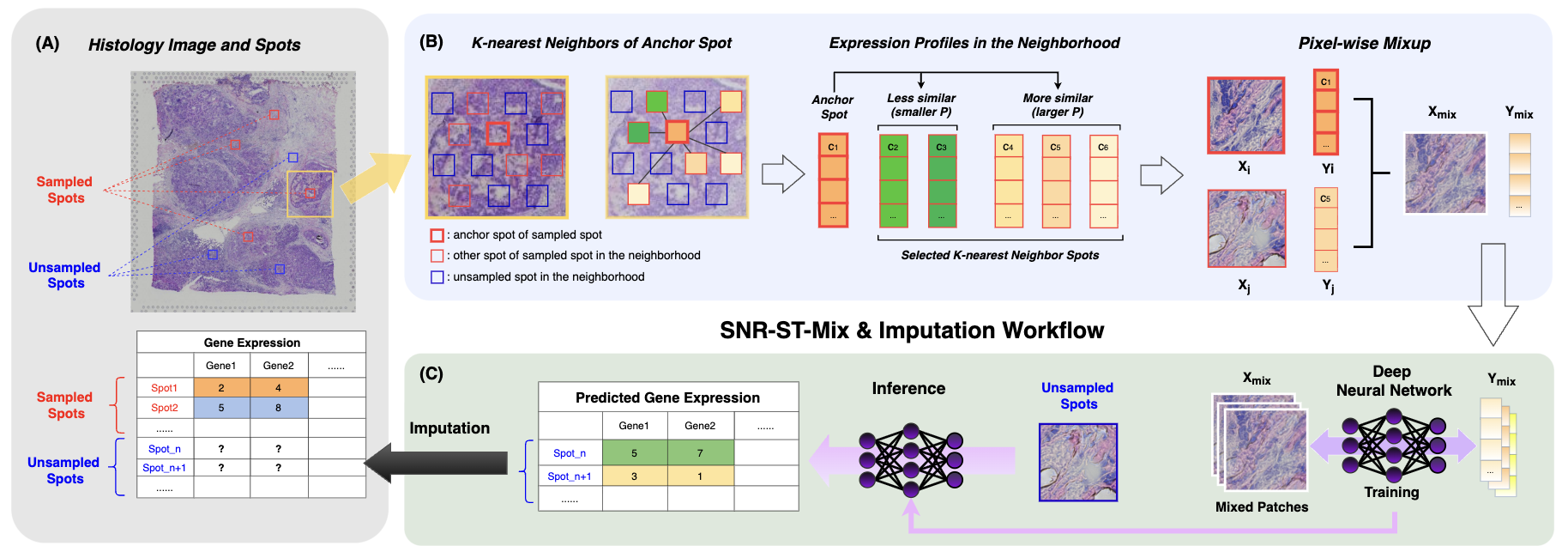}
    \caption{Overview of the SNR-ST-Mix framework. \textbf{(A)} Whole-slide images (WSIs) are divided into patches corresponding to ST spots; expression profiles are available for sampled spots, while unsampled spots lack expression measurements. \textbf{(B)} A KNN algorithm identifies each spot’s spatial neighborhood, with neighbors showing similar expression profiles assigned a higher mixup probability. \textbf{(C)} The model is trained on mixed patch–expression pairs to learn the relationship between histology and gene expression, then impute expression profiles for unsampled spots across the tissue.}
    \label{fig:framework}
\end{figure*}

SNR-ST-Mix contains two main components that contribute to the imputation of the expression profile. The method regularizes spot-level gene expression regression from histology with mixing sampled spots that are both (i) spatially proximate in tissue and (ii) similar in their target gene profiles. The first component utilizes a k-nearest neighbors (KNN) algorithm based on the spatial coordinates of the sampled ST data to preserve spatial smoothness. The second component defines a sampling probability that reflects the similarity between expression profiles of different spots, encouraging biologically consistent mixing. By training the model with mixed histology patches, SNR-ST-Mix reduces errors and leverages the strengths of deep neural networks for robust expression prediction. The complete workflow is illustrated in Figure~\ref{fig:framework}. 

\subsection{Spatial neighbors selection} 
\label{subsec:spatial_knn}
Consider a dataset of histology patches from a whole-slide image $\mathcal{D}=\{(x_i, y_i, c_i)\}_{i=1}^N$, where $x_i \in \mathbb{R}^{3 \times H \times W}$ is an H\&E patch centered at spot $i$, $y_i \in \mathbb{R}^{G}$ is the gene-expression vector for $G$ genes, and $c_i \in \mathbb{R}^2$ are spot coordinates. Spatial proximity is measured by Euclidean distance in coordinate space  
$ d_{ij} \;=\; \|c_i - c_j\|_2$.
The spatial information preserved in ST data reveals strong correlations among neighboring spots. This characteristic indicates that nearby spots tend to share similar cell-type compositions and microenvironmental features, leading to comparable patches across short spatial scales. For each anchor $i$, define its neighborhood of size $K$ by
\begin{equation}
\mathcal{N}_K(i) = \{\, j \ne i \mid d_{ij} \text{ is among the } K \text{ smallest distances} \,\}.
\label{eq:knn}
\end{equation}
Restricting mixup partner selection to $\mathcal{N}_K(i)$ implicitly encodes this local spatial dependency, ensuring that mixed pairs exhibit limited target discrepancy, while global mixing would blur tissue boundaries and gradients. As expression in ST data varies smoothly across tissue at the scale of spot coordinates, the spatial structure inside the ST data can be represented as
\begin{equation}
y_i \;=\; g^\star(c_i) + \varepsilon_i, 
\qquad 
\mathbb{E}[\varepsilon_i]=0,
\qquad
\text{Cov}(\varepsilon_i)=\Sigma_i,
\label{eq:lipschitz_prior}
\end{equation}
where $g^\star: \mathbb{R}^2 \rightarrow \mathbb{R}^G$ represents a latent spatial field that varies smoothly across tissue, and $\Sigma_i$ captures heteroscedastic noise. The spatial correlation can be modeled by a Lipschitz prior \citep{Zhou2024}, 
\begin{equation}
\|g^\star(c)-g^\star(c')\|_2 \le L_c \|c-c'\|_2.
\label{eq:lipschitz_prior}
\end{equation}
Under this prior, the expected discrepancy between expression profiles of neighboring spots $j \in \mathcal{N}_K(i)$ satisfies
\begin{equation}
\mathbb{E}\big[\|y_i - y_j\|_2^2 \,\big|\, c_i,c_j\big] 
\;\le\; L_c^2\, d_{ij}^2 \;+\; \operatorname{Tr}(\Sigma_i) + \operatorname{Tr}(\Sigma_j).
\label{eq:lipschitz_bound}
\end{equation}
Hence, selecting mixup partners from the $K$ nearest neighbors bounds the expected expression discrepancy by the spatial distance and noise term. This design introduces an explicit spatially local inductive bias, which stabilizes training and preserves tissue-specific structure. Moreover, spots within a neighborhood often show a morphological consistency, further reducing the noise introduced by mixing morphologically distinct regions.

\begin{algorithm}[t]
\small
\DontPrintSemicolon
\SetKwInput{Input}{Input}
\SetKwInput{Require}{Require}
\SetKwInput{Output}{Output}
\SetKwFunction{Zscore}{Z\!-\!score}
\SetKwFunction{KNN}{KNN}
\SetKwFunction{Norm}{Normalize}
\SetKwFunction{Cat}{Categorical}
\SetKwFunction{Beta}{Beta}
\SetKwFunction{Exp}{exp}
\SetKwProg{Proc}{Procedure}{}{}

\Input{Training set $\mathcal{D}=\{(x_i,y_i,c_i)\}_{i=1}^{N}$, $K$ (neighbors), label distance $d_{\text{label}}(\cdot,\cdot)$, bandwidth $\sigma$, mixup parameter $\alpha$, learning rate $\eta$}

\BlankLine
\textbf{(A) Precompute KNN and Sampling Probability}\;

\For{$i = 1$ \KwTo $N$}{
  $\mathcal{N}_K(i) \leftarrow$ \KNN{$c_i$, $K$} 
  
  \ForEach{$j \in \mathcal{N}_K(i)$}{
    $d_{\text{label}}(y_i,y_j) = \|y_i - y_j\|_2$\ \tcp*{\footnotesize label distance}
    $S_{ij} = \exp\!\Big(-\frac{d_{\text{label}}(y_i,y_j)}{2\sigma^2}\Big)$ \ \tcp*{\footnotesize Expression similarity}
    $P(j\mid i) \leftarrow \frac{S_{ij}}{\sum_{k\in\mathcal{N}(i)} S_{ik}}$ \tcp*[r]{\footnotesize Sampling Probability}
    }
}
\BlankLine

\textbf{(B) Training with SNR-ST-Mix}\;

Randomly initialize $\theta$\;

\While{not converged}{
  Sample a minibatch $\mathcal{B} \subset \{1,\ldots,N\}$\;
  
  $\mathbb{X}_{\text{mix}} \leftarrow \emptyset,\;\mathbb{Y}_{\text{mix}} \leftarrow \emptyset$\;
  
  \ForEach{$i \in \mathcal{B}$}{
     $j \sim \Cat\!\big(P(\cdot \mid i)\big)$ with $j \in \mathcal{N}(i)$\;
     
     $\lambda \sim \Beta(\alpha,\alpha)$\;
     
     $x_{\text{mix}} \leftarrow \lambda\,x_i + (1-\lambda)\,x_j$\;
     
     $y_{\text{mix}} \leftarrow \lambda\,y_i + (1-\lambda)\,y_j$\;
     
     Append $x_{\text{mix}}$ to $\mathbb{X}_{\text{mix}}$ and $y_{\text{mix}}$ to $\mathbb{Y}_{\text{mix}}$\;
  }
  $\hat{Y} \leftarrow g_\theta\!\big(f_\theta(\mathbb{X}_{\text{mix}})\big)$;
  $L \leftarrow \ell(\hat{Y}, \mathbb{Y}_{\text{mix}})$\;
  
  $\theta \leftarrow \theta - \eta \,\nabla_{\theta} L$
}
\caption{SNR-ST-Mix Algorithm}
\label{alg:SNR-ST}
\end{algorithm}

\subsection{Similar expression selection}
\label{subsec:label_similarity}
Given the anchor spot $i$ with gene vector $y_i \in \mathbb{R}^G$, SNR-ST-Mix assigns higher sampling probability to neighbors whose gene expression profiles are similar to $y_i$. Let $\mathcal{N}_K(i)$ denote the spatial $K$-nearest neighbors selected in Sec.~\ref{subsec:spatial_knn}. Within this neighborhood, expression similarity is measured in the local gene space using a Gaussian kernel:
\begin{equation}
\text{Sim}_{ij} \;=\; \exp\!\Big(-\frac{d_{\text{label}}(y_i,y_j)}{2\sigma^2}\Big),
\qquad j\in\mathcal{N}_K(i),
\label{eq:Gaussian_kernel}
\end{equation}
where $d_{\text{label}}(y_i,y_j) = \|y_i - y_j\|_2$, and $\sigma>0$ is a bandwidth controlling how rapidly similarity decays with expression discrepancy.

The similarities are normalized to form a conditional sampling distribution over the spatial neighbor set:
\begin{equation}
P_y(j \mid i) \;=\; \frac{\text{Sim}_{ij}}{\sum\limits_{k \in \mathcal{N}_K(i)} \text{Sim}_{ik}}, 
\quad j \in \mathcal{N}_k(i).
\label{eq:sampling_prob}
\end{equation}
In the neighborhood that preserves the spatial smoothness, this probabilistic weighting encourages mixup between spatially close spots with consistent expression patterns, avoiding interpolation between dissimilar or heterogeneous regions.

Mixup between the anchor spot $i$ and a pairing spot $j$ takes a linear interpolation on both the input H\&E patches and their corresponding expression profiles:
\begin{equation}
x_{\text{mix}} \;=\; \lambda x_i + (1-\lambda)x_j,
\quad
y_{\text{mix}} \;=\; \lambda y_i + (1-\lambda)y_j,
\label{eq:pixel_mixup}
\end{equation}
where the mixing ratio $\lambda\sim\mathrm{Beta}(\alpha,\alpha)$ with $\alpha>0$. As the inputs are histology image patches, mixup is performed on each pixel. As SNR-ST-Mix assigns sampling probability in~\eqref{eq:sampling_prob} based on label similarity of expression profiles, it effectively reduces the noise introduced by mixup. 
Under independent, zero-mean spot noise with covariances $\Sigma_i$ and $\Sigma_j$, the conditional covariance of the mixed expression is
\begin{equation}
\mathrm{Cov}(y_{mix} \mid i,j,\lambda)
=\lambda^2\Sigma_i+(1-\lambda)^2\Sigma_j
< \max\{\Sigma_i,\Sigma_j\},
\label{eq:var-reduce}
\end{equation}
for $\lambda\in(0,1)$. 
Label-similarity–weighted sampling reduces variance through noise averaging while maintaining small bias, since $|y_i-y_j|_2$ remains small under Eqs~\eqref{eq:Gaussian_kernel} and~\eqref{eq:sampling_prob}. This improves the bias--variance tradeoff especially under spotwise heteroscedasticity ($\Sigma_i\neq\Sigma_j$), which is common in ST.


By combining spatial locality with label-similarity weighting, SNR-ST-Mix performs a biologically consistent mixup that preserves local smoothness while mitigating noise from dissimilar regions. The overall procedure for sampling and training is summarized in Algorithm~\ref{alg:SNR-ST}.

\subsection{Loss functions}
\label{subsec:loss}
The predictor $f_\theta: \mathcal{X} \to \mathbb{R}^G$ is trained to predict gene expression for unsampled spots using the mixed samples created from spatial KNN partners with label-aware reweighting. We elaborate on the loss function for training with SNR-ST-Mix samples, which enhances the conventional vicinal risk minimization by incorporating two structure-aware regularizers and a mild correlation prior.

The mixed sample data $x_{mix}$ with $y_{mix}$ in Section~\ref{subsec:label_similarity} should contribute to the construction of the sample-specific expression profile predictor $f_\theta: \mathcal{X} \to \mathbb{R}^G$. The primary objective is mean squared error (MSE) on mixed pairs:
\begin{equation}
\mathcal{L}_{\text{vic}}(\theta)
\;=\;
\frac{1}{|\mathcal{B}|}\sum_{(i,j)\in\mathcal{B}}
\big\|f_\theta(x_{mix}) - y_{mix}\big\|_2^2 .
\label{eq:loss_vic}
\end{equation}
where spots $i$ and $j$ are spatially close and have similar labels.

As ST data exhibits structured inter-spot dependencies, $f_\theta$ is encouraged to be locally linear along the edge between $x_i$ and $x_j$. Therefore, a mixup consistency penalty is introduced:
\begin{equation}
\mathcal{L}_{\text{cons}}(\theta)
\;=\;
\frac{1}{|\mathcal{B}|}\sum_{(i,j)\in\mathcal{B}}
\Big\|f_\theta(x_{mix})
- \big(\lambda\,f_\theta(x_i) + (1-\lambda)\,f_\theta(x_j)\big)
\Big\|_2^2 ,
\label{eq:loss_cons}
\end{equation}
where $\lambda\in (0,1)$ is the mixup coefficient drawn from Beta $(\alpha, \alpha)$. This term matches the prediction on the mixed input to the mixture of endpoint predictions, explicitly controlling unnecessary curvature of $f_\theta$ along tissue-manifold directions. Thus, it stabilizes the predictor against small but ubiquitous morphological drifts across adjacent spots.

To respect biological boundaries while promoting tissue-wise smoothness, we further align predicted and true differences along the spatial graph. Let $w_{ij}\propto P_y(j \mid i)$ in \eqref{eq:sampling_prob} be the sampling weight of the edge $(i,j)$ combining spatial affinity and label similarity. We align prediction differences with target differences across sampled edges:
\begin{equation}
\mathcal{L}_{\text{edge}}(\theta)
\;=\;
\frac{1}{|\mathcal{B}|}\sum_{(i,j)\in\mathcal{B}}
w_{ij}
\big\|\underbrace{\big(f_\theta(x_i)-f_\theta(x_j)\big)}_{\Delta f_\theta(i,j)}
-
\underbrace{\big(y_i-y_j\big)}_{\Delta y(i,j)}
\big\|_2^2 .
\label{eq:loss_edge}
\end{equation}
When neighboring targets are similar, $\|\Delta y(i,j)\|_2$ is small. The dominant effect of~\eqref{eq:loss_edge} is to penalize $\sum_{(i,j)} w_{ij}\|\Delta f_\theta(i,j)\|_2^2$, which is the standard graph Laplacian smoothness penalty on predictions. In contrast, at genuine boundaries where $\|\Delta y(i,j)\|_2$ is large, the alignment form in~\eqref{eq:loss_edge} prevents oversmoothing by allowing correspondingly large prediction differences.

Besides, we include a mild correlation prior on the mixed pairs,
\begin{equation}
\mathcal{L}_{\mathrm{corr}}(\theta)
\;=\;
-\frac{1}{G}\sum_{g=1}^G
\mathrm{Pearson}\!\left(\, f_\theta(x_{mix})_{:,g},\; {y_{mix}}_{:,g}\,\right),
\label{eq:corr}
\end{equation}
which biases optimization toward higher Pearson correlation without destabilizing the MSE fit.

Combining the four components yields the final training objective
\begin{equation}
\mathcal{L}(\theta)
\;=\;
\mathcal{L}_{\mathrm{vic}}(\theta)
\;+\;
\lambda_{\mathrm{cons}}\,\mathcal{L}_{\mathrm{cons}}(\theta)
\;+\;
\lambda_{\mathrm{edge}}\,\mathcal{L}_{\mathrm{edge}}(\theta)
\;+\;
\lambda_{\mathrm{corr}}\,\mathcal{L}_{\mathrm{corr}}(\theta),
\label{eq:total}
\end{equation}
with nonnegative coefficients $\lambda_{\mathrm{cons}},\lambda_{\mathrm{edge}},\lambda_{\mathrm{corr}}$. This composite loss function reduces target variances while adding two principled regularizers for mixup consistency and Laplacian alignment.

\subsection{Datasets and data preprocessing}
\label{subsec:datasets}
We use 10x Genomics Visium datasets from the HEST-1K collection \citep{Jaume2024HEST1k}, which provides diverse spatial transcriptomics (ST) profiles paired with whole-slide histology images. Specifically, we select two human breast cancer samples(TENX13, TENX14), two bowel cancer samples from the colon molecular atlas (MISC72, MISC73), two human healthy heart samples (MISC128, MISC129), one ovarian cancer sample (TENX65), and one prostate cancer sample (TENX46). 

For each sample, the H\&E WSI is divided into local image patches corresponding to Visium capture spots based on spatial barcode coordinates (spot diameter = 55 µm). Each patch is resized to $224\times 224$ pixels and normalized to the ImageNet mean and standard deviation to match the input distribution expected by the pretrained ViT backbone. Gene expression counts are log-normalized and filtered by requiring each cell to express at least one gene and each gene to be detected in at least one cell. We select the top 250 genes with the highest mean expression across all spatial spots to reduce noise and focus on the most informative genes.

\subsection{Implementation details}
The training sets and validation sets are randomly selected from all spots with a proportion of 30$\%$ training spots, 20$\%$ validation spots, and 50$\%$ testing spots for all experiments. Gene expression imputation uses a Vision Transformer (ViT) backbone that has been pretrained on ImageNet. We modify ViT with the classification head removed and append a single linear projection to the $G$-dimensional gene space. The encoder is regularized with a dropout rate of $0.2$ before the prediction head. 

The entire network is fine-tuned from end-to-end with AdamW optimizer using the learning rate $3\times10^{-5}$, $\beta=(0.9,0.98)$, $\varepsilon=10^{-6}$, and weight decay $10^{-4}$. We adopt cosine annealing for the learning rate over the full training. We train for at most $200$ epochs with early stopping, and the best model based on validation MSE is used for evaluation in the untouched test set.
All experiments are conducted on an NVIDIA 4090 GPU using the PyTorch framework. 

The hyperparameter of mixup strength is set to $\alpha = 1.0$, so that $\lambda\sim\mathrm{Beta}(\alpha,\alpha)$ is approximately uniform on $[0,1]$, which avoids concentrating the mass near the endpoints and yields a balanced range of interpolation strength. The bandwidth $\sigma$ for the label-similarity kernel is set to $\sigma = \mathrm{median}\{{ \lVert y_i-y_j\rVert_2}\}/2$, which is one-half of the median pairwise label distance measured on the training set, thus neighbors closer than the median are substantially upweighted. The loss weights are set to $\lambda_{\mathrm{cons}}=0.5$, $\lambda_{\mathrm{edge}}=50$, and $\lambda_{\mathrm{corr}}=0.05$.

\subsection{Baselines and evaluation metrics}
We compare SNR-ST-Mix against the baseline and several representative augmentation methods, including basic augmentation, vanilla mixup, and CutMix. The baseline is a ViT–based regression model trained directly on histology patches without any data augmentation. The same ViT backbone was trained with standard image augmentations, including random horizontal or vertical flips and random cropping. It assesses whether simple geometric perturbations improve model robustness and generalization. We also assess the capabilities of Mixup \citep{Zhang2017} and CutMix \citep{Yun2019} across various ST datasets. While vanilla mixup interpolates patches based on pixels, CutMix is a region-wise augmentation that replaces a rectangular region from one patch with that from another, and linearly mixes their labels based on region area.

To evaluate the performance of the proposed method, we used three widely used metrics: mean squared error (MSE), mean absolute error (MAE), and Pearson correlation coefficient (PCC). They are calculated on the gene expression to assess the performance of the model trained on mixed samples.

\section{Experimental Results}

\begin{figure*}[htbp] 
    \centering
    \includegraphics[width=0.65\textwidth]{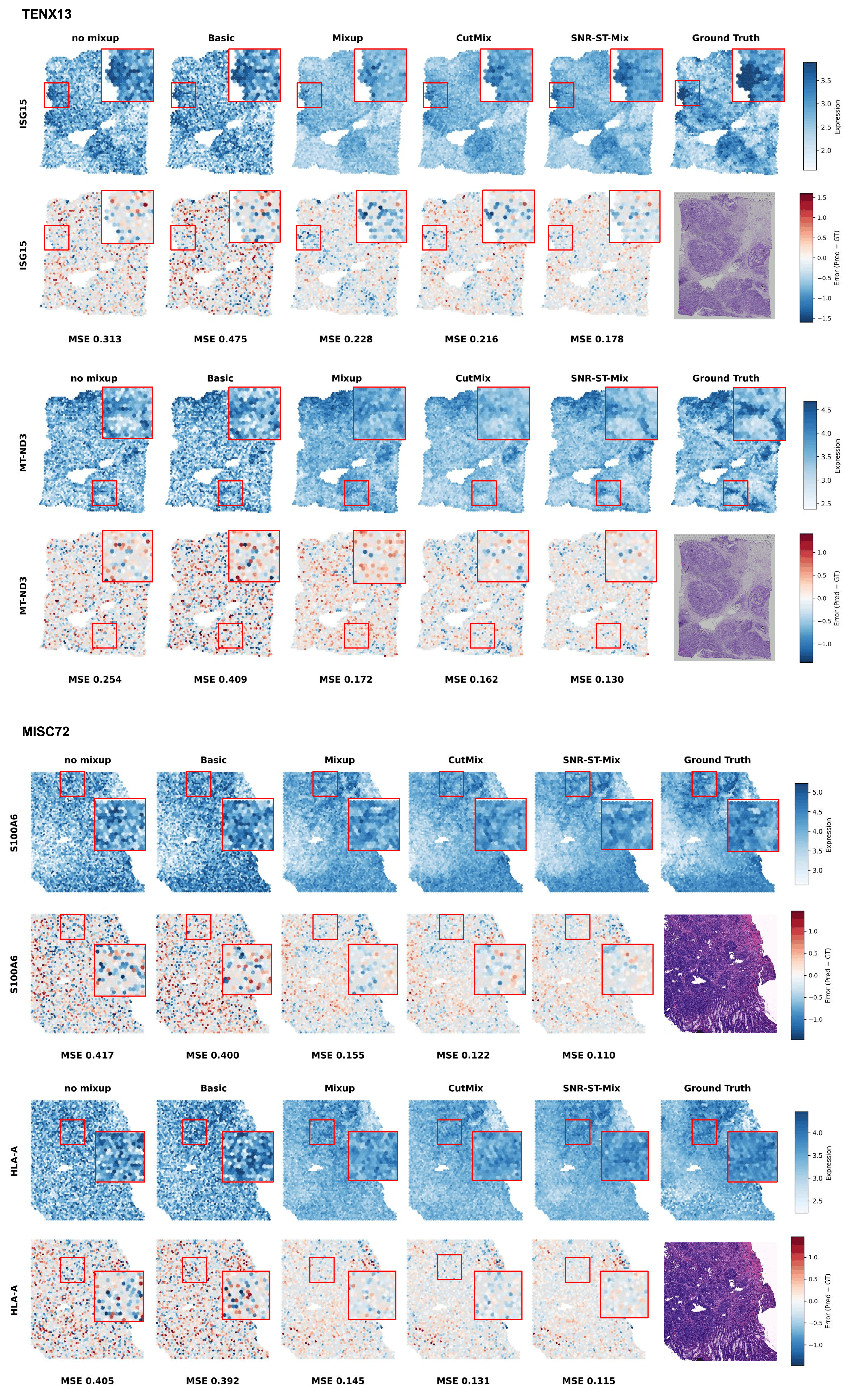}
    \caption{Visualization of predicted spatial gene expression maps for representative genes (ISG15, MT-ND3, S100A6, and HLA-A) from TENX13 and MISC72 datasets. Each row compares results from no mixup, basic augmentation, Mixup, CutMix, and SNR-ST-Mix against the ground truth. For each method, predicted expression maps are shown with error maps below. Red boxes indicate enlarged regions for detailed comparison, and the corresponding H\&E image of each tissue section is provided for spatial reference.}
    \label{fig:visualization}
\end{figure*}

To qualitatively assess the performance of SNR-ST-Mix compared with the baseline and existing augmentation strategies, we visualize predicted spatial expression maps for representative genes from TENX13 (breast cancer) and MISC72 (bowel cancer) datasets in Figure~\ref{fig:visualization}. Across all gene examples, Mixup-based methods better capture the overall spatial expression trends compared to the baseline and basic augmentations, demonstrating the advantage of interpolation-based augmentation in generalization. While all mixup-based methods reduce noise and encourage smoothness, SNR-ST-Mix produces expression patterns that most closely match the ground truth, with visibly sharper spatial boundaries and more coherent regional expression intensities.

\begin{figure*}[htb!] 
    \centering
    \includegraphics[width=0.95\textwidth]{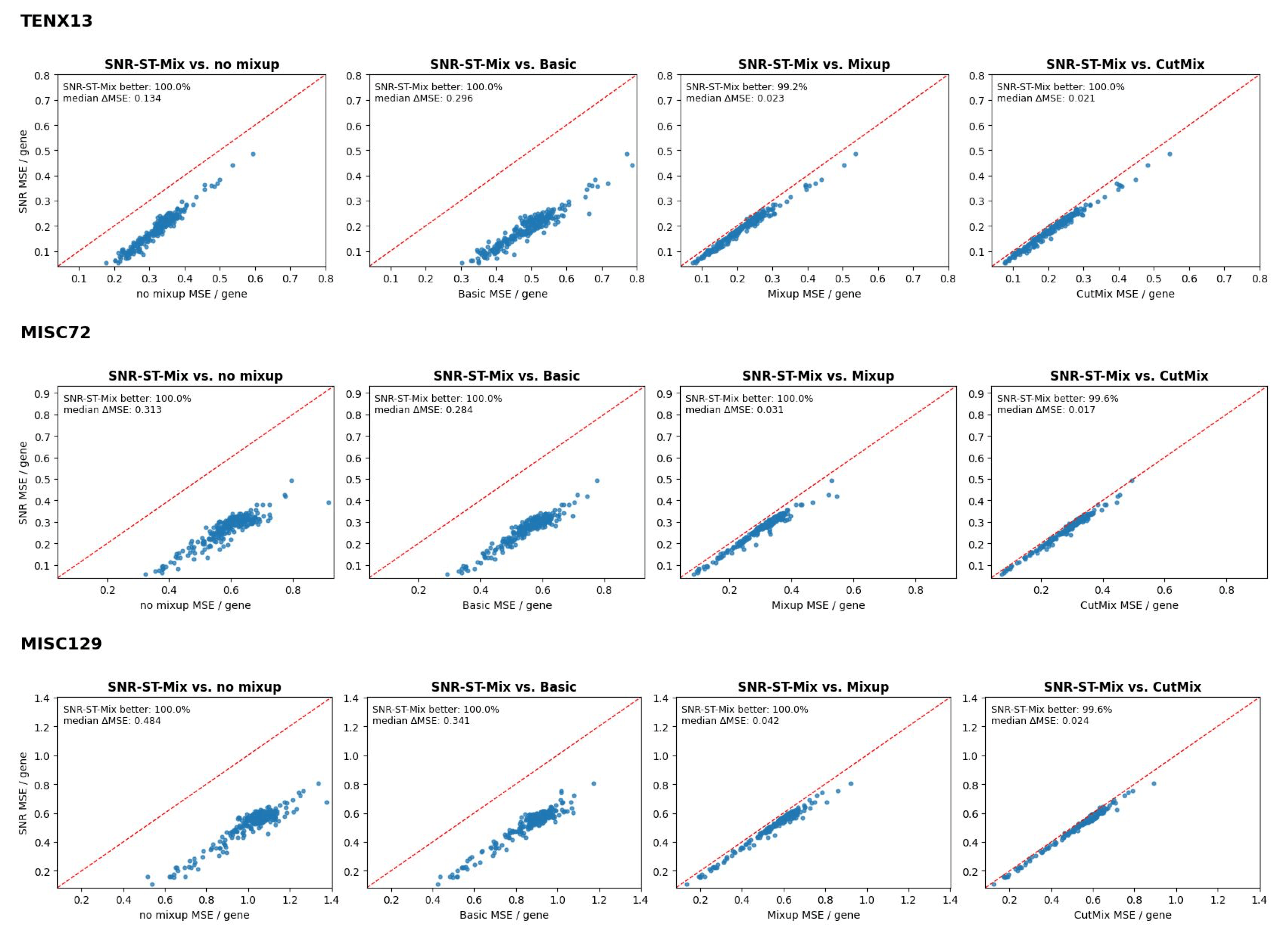}
    \caption{Comparison of per-gene MSE between SNR-ST-Mix and other augmentation methods for TENX13, MISC72, and MISC129. Each scatter plot compares the MSE values of individual genes under SNR-ST-Mix and one baseline method (no mixup, basic augmentation, Mixup, or CutMix). The red dashed line represents equal performance between methods, where points below the line indicate lower per-gene MSE for SNR-ST-Mix.}
    \label{fig:comparison_per_gene}
\end{figure*}

\begin{table}[htb!]
\fontsize{8}{10}\selectfont
\centering
\caption{Performance comparison of SNR-ST-Mix with baseline and augmentation methods across eight spatial transcriptomics datasets. The best performance for each dataset is highlighted in bold.}
\label{tab:performance}
\begin{tabular}{ll ccc}
\toprule
\cmidrule(lr){1-5}
\textbf{Dataset} & \textbf{Method} & \textbf{MSE ($\downarrow$)} & \textbf{MAE ($\downarrow$)} & \textbf{PCC ($\uparrow$)}\\

\midrule
\multirow{4}{*}{\shortstack{TENX13\\Breast\\Cancer}} 
& Baseline  & 0.2074 & 0.3081 & 0.4180 \\
& Basic Augmentation    & 0.2034 & 0.3127 & 0.4462 \\
& Mixup     & 0.2025 & 0.3083 & 0.4260\\
& CutMix    & 0.1978 & 0.3051 & 0.4415\\
& SNR-ST-Mix        & \bfseries 0.1886 & \bfseries 0.2911 & \bfseries 0.4748\\

\midrule
\multirow{4}{*}{\shortstack{TENX14\\Breast\\Cancer}} 
& Baseline          & 0.1718 & 0.2940 & 0.4229\\
& Basic Augmentation     & 0.1682 & 0.2906 & 0.4534 \\
& Mixup     & 0.1636 & 0.2852 & 0.4495\\
& CutMix    & 0.1594 & 0.2798 & 0.4573\\
& SNR-ST-Mix        & \bfseries 0.1463 & \bfseries 0.2710 & \bfseries 0.4802\\

\midrule
\multirow{4}{*}{\shortstack{MISC72\\Bowel\\Cancer}} 
& Baseline          & 0.2895 & 0.3789 & 0.4728\\
& Basic Augmentation     & 0.2889 & 0.3670 & 0.4755\\
& Mixup     & 0.2872 & 0.3698 & 0.4710\\
& CutMix    & 0.2820 & 0.3657 & 0.4781\\ 
& SNR-ST-Mix        & \bfseries 0.2542 & \bfseries 0.3436 & \bfseries 0.4925\\

\midrule
\multirow{4}{*}{\shortstack{MISC73\\Bowel\\Cancer}} 
& Baseline          & 0.2456 & 0.3510 & 0.4798 \\
& Basic Augmentation     & 0.2417 & 0.3437 & 0.4934 \\
& Mixup     & 0.2398 & 0.3391 & 0.4817\\
& CutMix    & 0.2354 & 0.3371 & 0.4924\\ 
& SNR-ST-Mix        & \bfseries 0.2227 & \bfseries 0.3254 & \bfseries 0.5064\\

\midrule
\multirow{4}{*}{\shortstack{TENX65\\Ovary\\Cancer}} 
& Baseline          & 0.2210 & 0.3368 & 0.5182 \\
& Basic Augmentation     & 0.2173 & 0.3346 & 0.5378 \\
& Mixup     & 0.2142 & 0.3286 & 0.5347\\
& CutMix    & 0.2090 & 0.3233 & 0.5514\\ 
& SNR-ST-Mix        & \bfseries 0.2045 & \bfseries 0.3222 & \bfseries 0.5568 \\

\midrule
\multirow{4}{*}{\shortstack{TENX46\\Prostate\\Cancer}} 
& Baseline          & 0.1617 & 0.2990 & 0.3511 \\
& Basic Augmentation     & 0.1521 & 0.2916 & 0.3695 \\
& Mixup     & 0.1526 & 0.2910 & 0.3460\\
& CutMix    & 0.1500 & 0.2886 & 0.3643\\ 
& SNR-ST-Mix        & \bfseries 0.1456 & \bfseries 0.2851 & \bfseries 0.3684\\

\midrule
\multirow{4}{*}{\shortstack{MISC128\\Heart\\Healthy}} 
& Baseline          & 0.5087 & 0.5106 & 0.4427\\
& Basic Augmentation     & 0.4971 & 0.4983 & 0.4383\\
& Mixup     & 0.4861 & 0.4991 & 0.4404\\
& CutMix    & 0.4768 & 0.4896 & 0.4515\\ 
& SNR-ST-Mix        & \bfseries 0.4734 & \bfseries 0.4867 & \bfseries 0.4559 \\

\midrule
\multirow{4}{*}{\shortstack{MISC129\\Heart\\Healthy}} 
& no mixup          & 0.5659 & 0.5523 & 0.4067\\
& Basic Augmentation     & 0.5450 & 0.5411 & 0.4208\\
& Mixup     & 0.5317 & 0.5362 & 0.4258\\
& CutMix    & 0.5219 & 0.5275 & 0.4352\\ 
& SNR-ST-Mix        & \bfseries 0.5163 & \bfseries 0.5268 & \bfseries 0.4385\\

\bottomrule
\end{tabular}
\end{table}

In the TENX13 dataset, the predicted expression of ISG15 generated by SNR-ST-Mix recovers the high-expression region in the left-middle portion of the sample more accurately. Compared to Mixup and CutMix, SNR-ST-Mix produces a clearer and more intense signal that aligns closely with the ground-truth distribution. It also markedly reduces noise, lowering the MSE from 0.313 in the no-mixup baseline and 0.475 in basic augmentation to 0.178. Similarly, for MT-ND3, SNR-ST-Mix captures the sharp boundary in the bottom region, preserving the structure of the high-expression area while effectively suppressing noise in low-expression regions. The lowest MSE of 0.130, compared to 0.172 with Mixup and 0.162 with CutMix, demonstrates that SNR-ST-Mix performs better in modeling morphology-dependent variations of gene expression in histology images. In the MISC72 sample, SNR-ST-Mix also produces smoother and more spatially consistent predictions for S100A6 and HLA-A. Although the ground truth does not exhibit obvious variations in the expression map, SNR-ST-Mix still achieves the lowest MSE among all mixup-based methods. This finding suggests that SNR-ST-Mix not only performs well in tissues with sharp spatial expression gradients but also maintains robustness in samples with relatively homogeneous expression patterns. 

Beyond the representative gene examples visualized, the superior performance of SNR-ST-Mix is observed in nearly all genes across datasets. Figure~\ref{fig:comparison_per_gene} presents per-gene MSE comparisons between SNR-ST-Mix and other augmentation strategies in TENX13 (breast cancer), MISC72 (bowel cancer), and MISC129 (healthy heart). In all cases, almost every points lie below the red diagonal line, indicating that SNR-ST-Mix achieves lower prediction errors for the vast majority of genes. In TENX13, 100\% of genes perform better under SNR-ST-Mix compared to the no-mixup, basic, and CutMix methods, and 99.2\% outperform Mixup, with a median MSE difference ranging from 0.296 to 0.021. The median MSE differences with the no-mixup baseline and basic augmentation are large (0.134 and 0.296), and the difference with mixup-based methods is relatively smaller (0.023 for Mixup and 0.021 for CutMix). A similar pattern is observed in MISC72, where over 99.6\% of genes show improved accuracy, achieving median MSE difference values of 0.313, 0.284, 0.031, and 0.017 relative to the four respective baselines. Even in the more sparse and challenging dataset of MISC129, SNR-ST-Mix consistently improves predictions for over 99.6\% of genes, with a median MSE difference ranging from 0.484 to 0.024.

\begin{table}[htb!]
\fontsize{8}{10}
\centering
\footnotesize
\caption{Ablation study of SNR-ST-Mix showing the effect of each component: spatial neighborhood constraint (KNN), label similarity weighting, and loss design. Each component is added incrementally to vanilla Mixup.}
\label{tab:ablation}
\setlength{\tabcolsep}{3pt}
\begin{tabular}{
  l c c c
  S[table-format=1.4] S[table-format=1.4] S[table-format=1.4]
}
\toprule
\textbf{Dataset} & \textbf{KNN} & \textbf{Label} & \textbf{Loss} &
\textbf{MSE ($\downarrow$)} & \textbf{MAE ($\downarrow$)} & \textbf{PCC ($\uparrow$)} 
\\
\midrule

\multirow{5}{*}{TENX13}
& \xmark & \xmark & \xmark & 0.2025 & 0.3083 & 0.4260 \\
& \cmark & \xmark & \xmark & 0.1984 & 0.3048 & 0.4363 \\
& \xmark & \cmark & \xmark & 0.1989 & 0.3021 & 0.4365 \\
& \cmark & \cmark & \xmark & 0.1944 & 0.2973 & 0.4564 \\
& \cmark & \cmark & \cmark & 0.1886 & 0.2911 & 0.4748 \\
\midrule

\multirow{5}{*}{TENX14}
& \xmark & \xmark & \xmark & 0.1636 & 0.2852 & 0.4495 \\
& \cmark & \xmark & \xmark & 0.1607 & 0.2819 & 0.4551 \\
& \xmark & \cmark & \xmark & 0.1627 & 0.2845 & 0.4514 \\
& \cmark & \cmark & \xmark & 0.1553 & 0.2776 & 0.4648 \\
& \cmark & \cmark & \cmark & 0.1463 & 0.2710 & 0.4802 \\ 
\midrule

\multirow{5}{*}{MISC72}
& \xmark & \xmark & \xmark & 0.2872 & 0.3698 & 0.4710 \\
& \cmark & \xmark & \xmark & 0.2853 & 0.3670 & 0.4765 \\
& \xmark & \cmark & \xmark & 0.2865 & 0.3679 & 0.4736 \\
& \cmark & \cmark & \xmark & 0.2774 & 0.3603 & 0.4867 \\
& \cmark & \cmark & \cmark & 0.2542 & 0.3436 & 0.4925 \\
\midrule

\multirow{5}{*}{MISC73}
& \xmark & \xmark & \xmark & 0.2398 & 0.3391 & 0.4817 \\
& \cmark & \xmark & \xmark & 0.2389 & 0.3415 & 0.4910 \\
& \xmark & \cmark & \xmark & 0.2394 & 0.3383 & 0.4842 \\
& \cmark & \cmark & \xmark & 0.2324 & 0.3331 & 0.4966 \\
& \cmark & \cmark & \cmark & 0.2227 & 0.3254 & 0.5064 \\
\bottomrule
\end{tabular}
\end{table}

Table~\ref{tab:performance} summarizes the quantitative comparison of SNR-ST-Mix with baseline and existing augmentation strategies across eight spatial transcriptomics datasets. SNR-ST-Mix consistently achieves the lowest MSE and MAE, indicating better imputation accuracy compared with all approaches. A higher PCC of SNR-ST-Mix across all datasets also reflects improved preservation of the linear relationship between predicted and true gene expression profiles.  
In the breast cancer datasets (TENX13 and TENX14), SNR-ST-Mix reduces the MSE from 0.2074 and 0.1718 at baseline to 0.1886 and 0.1491, achieving the highest PCC values of 0.4711 and 0.4752. In the bowel cancer datasets (MISC72 and MISC73), SNR-ST-Mix also outperforms all other methods in error reduction and correlation improvement. While traditional mixup-based augmentations of Mixup and CutMix produce moderate gains over the baseline, SNR-ST-Mix achieves a notably lower MSE of 0.2542 in MISC72, compared with 0.2872 in Mixup and 0.2820 in CutMix. 
As SNR-ST-Mix improves prediction accuracy across all datasets, the magnitude of improvement varies among tissues and organs. In the ovary (TENX65) and prostate (TENX46) datasets, SNR-ST-Mix surpassed the baseline but achieved results comparable to CutMix, suggesting that in structurally homogeneous tissues with lower biological variability, the benefits of adaptive neighborhood mixing are less pronounced yet still stabilizing. For healthy heart tissues (MISC128 and MISC129), which are inherently more challenging due to sparse expression signals, SNR-ST-Mix remains competitive and achieves the lowest MSE of 0.4734 and 0.5163, compared to 0.5087 and 0.5669 from the baselines, while maintaining superior correlation values.

\begin{figure*}[htb!] 
    \centering
    \includegraphics[width=0.99\textwidth]{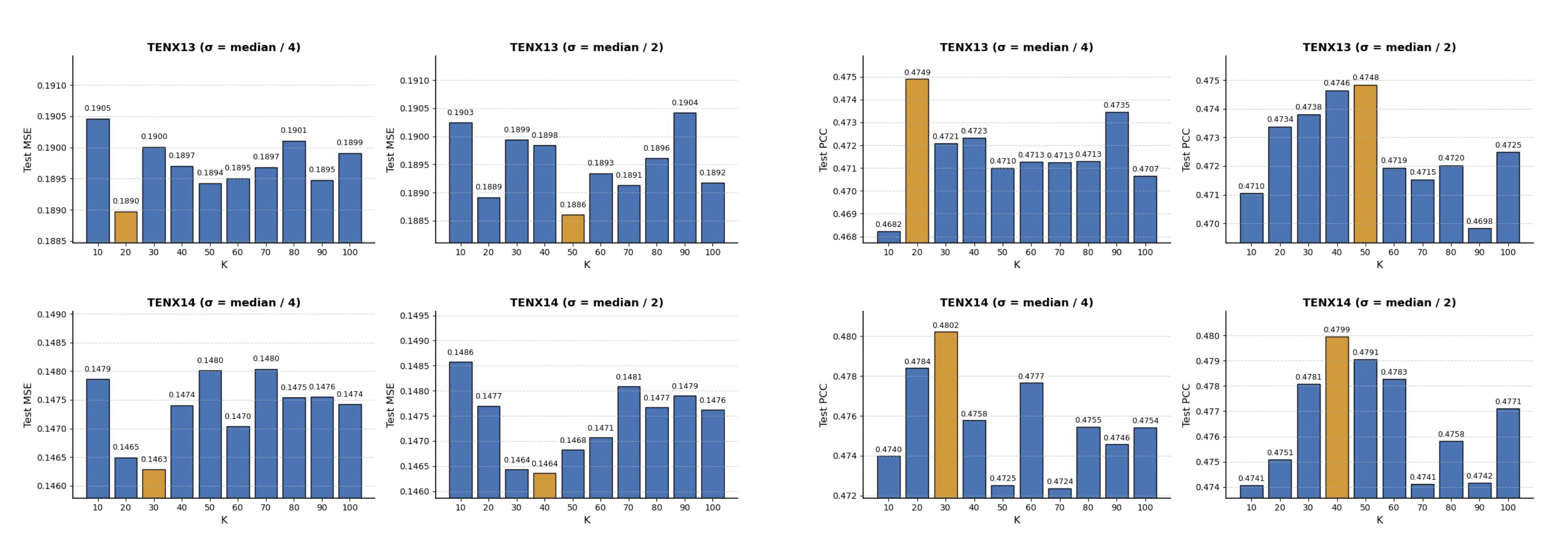}
    \caption{Ablation study on the effect of hyperparameter settings for the number of spatial neighbors $K$ and label-similarity bandwidth $\sigma$ in the TENX13 and TENX14 datasets. Test MSE (left) and PCC (right) are plotted for varying numbers of spatial neighbors ($K$) under two label-similarity bandwidths, $\sigma = \mathrm{median} /4$ and $\sigma = \mathrm{median} /2$. Orange bars denote the best results for each configuration. }
    \label{fig:K_selection}
\end{figure*}

To evaluate the contribution of each component in SNR-ST-Mix, we conducted ablation studies where KNN, label similarity weighting, and loss design were progressively integrated into the vanilla Mixup method, as shown in Table~\ref{tab:ablation}. While introducing either KNN or label similarity individually in mixup candidate selection contributes to the augmented samples, the improvement becomes more pronounced when they are incorporated together. For example, in TENX13, adding kNN reduces MSE from 0.2025 to 0.1984, and label similarity achieves a comparable reduction to 0.1989. When KNN and label similarity are combined, MSE decreases further to 0.1944, and PCC rises to 0.4564, demonstrating a synergistic effect between spatial and expression-based constraints. Similarly, in MISC72, KNN reduces MSE from 0.2872 to 0.2853, and label similarity reduces MSE slightly to 0.2865. When incorporating both, MSE reduces significantly to 0.2774 with a PCC of 0.4867 compared to 0.4710 at the baseline. This observation demonstrates that both KNN and label similarity are essential and complement each other in SNR-ST-Mix imputation. The PCC value in Table~\ref{tab:ablation} also demonstrates this finding. In TENX13, incorporating KNN or label similarity individually increases PCC from 0.4260 to 0.4363 and 0.4365, while combining both of them boosts PCC significantly to 0.4564.

As Table~\ref{tab:ablation} shows, our loss design also enhances performance; we further examine each loss component in SNR-ST-Mix. The results indicate that the consistency loss, edge loss, and Pearson loss all contribute positively to performance by providing effective regularization. In TENX13, introducing the consistency loss improves prediction accuracy and reduces MSE from 0.1944 to 0.1908. Adding the edge loss further decreases MSE to 0.1926, demonstrating the benefit of aligning prediction differences with true expression differences between spatial neighbors. Finally, incorporating a mild correlation loss leads to a higher PCC value of 0.4748. Together, these results confirm that each component of the composite loss serves a useful function in the expression imputation.

\begin{table}[htb!]
\centering
\small
\caption{Ablation study of loss components in SNR-ST-Mix for the TENX13 and TENX14 datasets. The table reports results when the mixup consistency loss ($\mathcal{L}_{\mathrm{cons}}$), edge loss ($\mathcal{L}_{\mathrm{edge}}$), and correlation loss ($\mathcal{L}_{\mathrm{corr}}$) are added incrementally to MSE loss.}
\label{tab:ablation_loss}
\setlength{\tabcolsep}{3pt}
\begin{tabular}{
  l c c c c 
  S[table-format=1.4] S[table-format=1.4] S[table-format=1.4]
}
\toprule
\textbf{Dataset} & 
\textbf{$\mathcal{L}_{\mathrm{cons}}$} & \textbf{$\mathcal{L}_{\mathrm{edge}}$} & \textbf{$\mathcal{L}_{\mathrm{corr}}$} & 
\textbf{MSE ($\downarrow$)} & \textbf{MAE ($\downarrow$)} & \textbf{PCC ($\uparrow$)}
\\
\midrule
\multirow{5}{*}{TENX13}
& \xmark & \xmark & \xmark & 0.1944 & 0.2973 & 0.4564 \\
& \cmark & \xmark & \xmark & 0.1908 & 0.2933 & 0.4635 \\
& \xmark & \cmark & \xmark & 0.1926 & 0.2947 & 0.4597 \\
& \cmark & \cmark & \xmark & 0.1888 & 0.2913 & 0.4712 \\
& \cmark & \cmark & \cmark & 0.1886 & 0.2911 & 0.4748 \\
\midrule

\multirow{5}{*}{TENX14}
& \xmark & \xmark & \xmark  & 0.1553 & 0.2776 & 0.4648 \\
& \cmark & \xmark & \xmark  & 0.1474 & 0.2712 & 0.4776 \\
& \xmark & \cmark & \xmark  & 0.1477 & 0.2713 & 0.4705 \\
& \cmark & \cmark & \xmark  & 0.1463 & 0.2709 & 0.4797 \\
& \cmark & \cmark & \cmark  & 0.1463 & 0.2710 & 0.4802 \\

\bottomrule
\end{tabular}
\end{table}

To evaluate the influence of hyperparameters in SNR-ST-Mix, we assessed model performance under varying numbers of spatial neighbors ($K$) and bandwidth parameter ($\sigma$) used for label similarity weighting in Figure~\ref{fig:K_selection}. Across both TENX13 and TENX14 samples, the overall results show that SNR-ST-Mix is insensitive to hyperparameter variation, achieving consistently low MSE and high PCC across a broad range of settings. In TENX13, MSE values remain within a relatively narrow range from 0.1886 to 0.1905 for a wide range of $K \in [10, 100]$ neighbors considered in mixup. Likewise, PCC fluctuates minimally around 0.469 to 0.475, confirming that the method generalizes robustly without requiring precise tuning. The bandwidth parameter $\sigma$ is adaptively defined based on the median of pairwise label distances, and both $\sigma = \mathrm{median} /4$ and $\sigma = \mathrm{median} /2$ achieve comparable performance. In TENX14, the best MSE for different $\sigma$ are 0.1464 and 0.1463, with the corresponding PCC of 0.4802 and 0.4799. These results indicate that SNR-ST-Mix maintains stable performance across a wide range of neighborhood and similarity bandwidth parameters, highlighting its robustness and ease of application to diverse ST datasets.

\section{Discussion}
The experimental evidence presented in Figure~\ref{fig:visualization}--~\ref{fig:K_selection} and Table~\ref{tab:performance}--~\ref{tab:ablation_loss} demonstrates that SNR-ST-Mix provides substantial and consistent improvements in spatial gene expression prediction by producing biologically meaningful augmentations that consider both spatial and molecular structure. The visualization results (Figure~\ref{fig:visualization}) confirm that SNR-ST-Mix effectively preserves spatial context while reducing noise, leading to clearer and more accurate expression imputation across diverse tissue architectures. The consistently superior performance in quantitative comparisons (Table~\ref{tab:performance}) emphasizes the contribution of incorporating neighborhood information and expression similarity when mixing histology patches. By jointly considering the spatial smoothness and the molecular relationship between spots, SNR-ST-Mix generates augmented examples that maintain biologically consistent spatial patterns while enhancing model generalization. 

The ablation study (Table~\ref{tab:ablation}) further elucidates the contribution of each module within SNR-ST-Mix. The observed improvement when combining the KNN constraints and label-similarity weighting can be attributed to the complementary nature of spatial and transcriptomic relationships in ST data. Spatial proximity and transcriptomic similarity are related but not equivalent. Neighboring spots often share structural or microenvironmental features, but they can differ markedly in molecular composition, especially at tissue boundaries or in heterogeneous tumor regions. When only the KNN constraint is applied, mixup may preserve geometric continuity but risk blending biologically distinct regions. Conversely, using only label similarity encourages mixing between transcriptionally coherent spots, even when these spots are spatially distant, thereby ignoring critical histological context. By integrating both constraints, SNR-ST-Mix performs mixup between samples that are simultaneously close in physical space and coherent in expression space, producing augmented examples that more accurately reflect local biological variability within tissue-level structure.

While SNR-ST-Mix demonstrates its advantages, there are still some limitations for future investigation and extension. One potential limitation of the present framework is that its effectiveness remains partly sample-dependent. While SNR-ST-Mix provides consistent improvements across all datasets, the magnitude of these gains varies by tissue type, being larger in heterogeneous tumor samples and more modest in relatively homogeneous tissues, such as the healthy heart. This suggests that the optimal augmentation strength and neighborhood structure may depend on the intrinsic spatial variability of each tissue. Currently, the method uses a fixed neighborhood radius, which may not fully capture differences in local heterogeneity. Future work could introduce adaptive or learnable mechanisms that automatically adjust spatial neighborhoods or expression similarity thresholds based on the underlying tissue structure, enabling the model to tailor its augmentation behavior to each sample. Moreover, if additional sections or replicates from the same samples are available, SNR-ST-Mix could be further refined through sample-specific finetuning, potentially enhancing performance by adapting to the tissue characteristics \citep{Sikaroudi2023OOD}. Second, although the present study focuses on a pixel-level implementation for the regression task of expression imputation, the underlying principles of spatial-neighborhood constraint and label-similarity weighting are broadly applicable for mixup. These ideas can naturally extend to augmentation paradigms, such as CutMix, PuzzleMix, or other region-level mixing strategies for image classification tasks \citep{Yun2019, Kim2020,Zhang2025CellMix}. Incorporating biological structure into these frameworks could potentially enable patch selection and label mixing to respect local tissue organization rather than relying on random cut-and-paste operations. Future work may also include cell morphology in region-level mixup, where the mixed regions are guided by histological boundaries. Exploring these directions could lead to a broader class of biologically informed augmentation techniques that enhance generalization across diverse spatial omics and histopathology applications. Lastly, the Visium datasets used in this study provide gene expression aggregated over multiple cells, which can obscure sharp biological boundaries and limit the granularity of spatial relationships captured by the model \citep{Yang2025Spotiphy}. Since SNR-ST-Mix already relies on spatial proximity and expression similarity, its benefits would likely be amplified in single-cell or subcellular-resolution datasets like MERFISH, Xenium, or CosMx. At these resolutions, local neighborhoods truly represent individual cells or microenvironments, allowing the model’s KNN and label-similarity modules to operate on biologically meaningful scales. Alternatively, integrating our approach with deconvolution frameworks that infer cell-type-specific expression at the spot level could further improve its biological precision.

\section{Conclusion}

We introduce SNR-ST-Mix, the first regression-based mixup framework for improving gene expression prediction in spatial transcriptomics. By integrating spatial proximity and expression similarity into the augmentation process, the method generates interpolations that preserve local tissue structure, addressing the drawbacks of applying conventional mixup strategies in ST data. Comprehensive experiments demonstrate that SNR-ST-Mix consistently enhances imputation accuracy, stabilizes learning under data-limited conditions, and improves generalization without adding architectural complexity or computational burden. These finds emphasize the importance of biologically informed augmentation and highlight SNR-ST-Mix as a useful tool for advancing spatial transcriptomic analysis. As spatial technologies continue to evolve toward higher resolution and richer multimodal measurements, the principles introduced here offer a foundation for developing next-generation augmentation frameworks that more faithfully capture cellular and tissue-level organization.

\section{Code and Data Availability}
The code used in this study is available on the AIMP-Lab GitHub repository at \url{https://github.com/Advanced-AI-in-Medicine-and-Physics-Lab/SNR-ST-Mix}. All data used in this work are publicly available from the HEST dataset hosted on Hugging Face at \url{https://huggingface.co/datasets/MahmoodLab/hest/}.

\section{Disclosures}
The authors declare that there are no financial interests, commercial affiliations, or other potential conflicts of interest that could have influenced the objectivity of this research or the writing of this paper.

\bibliographystyle{spiejour}
\bibliography{report}

\end{spacing}
\end{document}